\documentclass[11pt,a4paper]{article}
\usepackage[utf8]{inputenc}

\usepackage[hyperref]{acl2019}
\usepackage{times}
\usepackage{latexsym}
\usepackage{graphicx} 
\usepackage{multirow}
\usepackage{array}
\usepackage{arydshln} 
\usepackage{amssymb,latexsym,pifont}
\usepackage{amsmath}
\usepackage{covington}
\usepackage{enumitem}
\usepackage{url}
\usepackage[normalem]{ulem} 
\usepackage{microtype}
\usepackage[absolute]{textpos}

\newcommand\Tstrut{\rule{0pt}{2.3ex}}       
\newcommand\TTstrut{\rule{0pt}{3.2ex}}       
\newcommand\Bstrut{\rule[-0.9ex]{0pt}{0pt}} 
\newcommand\BBstrut{\rule[-2.5ex]{0pt}{0pt}} 

\definecolor{darkblue}{rgb}{0.2, 0.2, 0.6}
\definecolor{midviolet}{rgb}{0.69, 0.0, 0.69}
\definecolor{maxmagenta}{rgb}{1.0, 0.0, 1.0}
\definecolor{darkgreen}{rgb}{0.0, 0.5, 0.0}

\def\semerr#1{\textcolor{red}{\uuline{\emph{#1}}}}
\def\fluency#1{\textcolor{blue}{\dotuline{#1}}}
\def\lexic#1{\textcolor{brown}{\dashuline{#1}}}
\def\inapstr#1{\textcolor{purple}{\uwave{#1}}}
\def\repeat#1{\textcolor{maxmagenta}{\sout{#1}}}

\newcommand{\g}[2]{\begin{tabular}[t]{@{}c@{}}#1\\#2\end{tabular}}
\newcommand{\stack}[1]{\begin{tabular}[t]{@{}c@{}}#1\end{tabular}}

\def\ODdel#1{\bgroup\markoverwith{\textcolor{darkgreen}{\rule[0.5ex]{2pt}{1pt}}}\ULon{#1}}

\definecolor{fjred}{rgb}{0.5, 0.0, 0.0}

\aclfinalcopy 
\def\aclpaperid{68} 

\title{Neural Generation for Czech: Data and Baselines}

\author{Ondřej Dušek \and Filip Jurčíček \\
  Charles University, Faculty of Mathematics and Physics \\
  Institute of Formal and Applied Linguistics \\
  Prague, Czech Republic \\
  \texttt{\{odusek,jurcicek\}@ufal.mff.cuni.cz} \\}

\date{}

\begin{document}
\maketitle

\begin{textblock*}{\textwidth}(2.5cm,1cm)
In \emph{Proceedings of INLG}, Tokyo, Japan, October 2019.
\end{textblock*}

\begin{abstract}
We present the first dataset targeted at end-to-end NLG in Czech in the restaurant domain, along with several strong baseline models using the sequence-to-sequence approach. 
While non-English NLG is under-explored in general, Czech, as a morphologically rich language, makes the task even harder: Since Czech requires inflecting named entities, delexicalization or copy mechanisms do not work out-of-the-box and lexicalizing the generated outputs is non-trivial. 

In our experiments, we present two different approaches to this this problem: (1) using a neural language model to select the correct inflected form while lexicalizing, (2) a two-step generation setup: our sequence-to-sequence model generates an interleaved sequence of lemmas and morphological tags, which are then inflected by a morphological generator.
\end{abstract}

\section{Introduction}
\label{sec:intro}

While most current neural NLG systems do not explicitly contain lan\-guage-specific components and are thus capable of multilingual generation in principle, there has been little work to test these capabilities experimentally.
This goes hand in hand with the scarcity of non-English training datasets for NLG -- the only data-to-text NLG set known to us is a small sportscasting Korean dataset \cite{chen_training_2010},\footnote{\url{http://www.cs.utexas.edu/users/ml/clamp/sportscasting/}} 
which only contains a limited number of named entities, reducing the need for their inflection.

\begin{figure*}[tb]
\centering
\includegraphics[width=\textwidth]{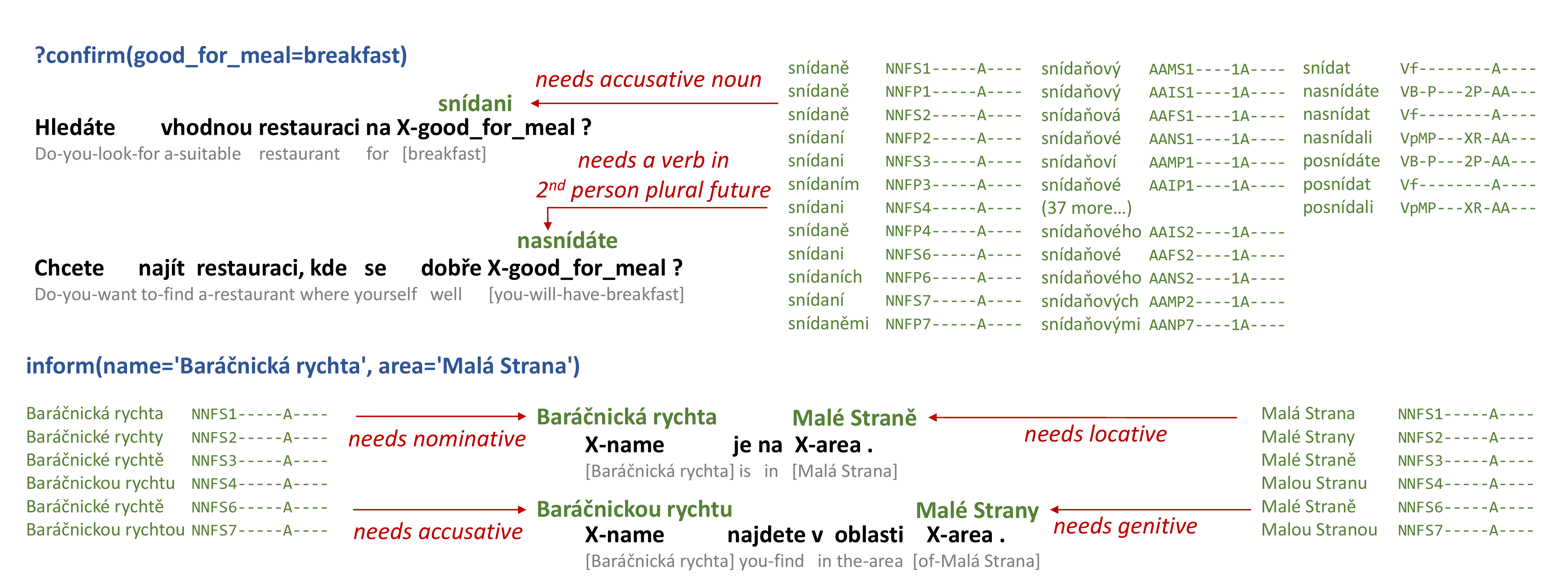}
\caption{Example of delexicalized generation in Czech. Input MRs are shown in bold blue, corresponding target (delexicalized) outputs in bold black, with ``X-'' marking slot value placeholders. English glosses are shown below each word in gray. Appropriate inflected forms to be filled into slot placeholders are shown in bold green, with lists of all possible forms along with their morphological tags \cite{hajic_disambiguation_2004}. Note that the surface form for ``X-good\_for\_meal'' can even have different parts-of-speech (left column: noun, middle: adjective, right: verb forms).}
\label{fig:lexic}
\end{figure*}

Since most generators are only tested on English, they do not need to handle grammar complexities not present in English. 
A prime example is the delexicalization technique used by most current generators \cite[e.g.,][]{oh_stochastic_2000,mairesse_phrase-based_2010,wen_stochastic_2015,wen_semantically_2015,juraska_deep_2018}: It is generally assumed that attribute (slot) values from the input meaning representation (MR) can be replaced by placeholders during generation and inserted into the output verbatim. 
Delexicalization or an analogous technique, such as a copy mechanism \cite{gu_incorporating_2016,gehrmann_end--end_2018}, is required for most generation scenarios to allow generalization to unseen entity names: sets of entities are open (potentially infinite and subject to change) while training data is scarce.
However, the verbatim insertion assumption does not hold for languages with extensive noun inflection -- attribute values need to be inflected here to produce fluent outputs (see Figure~\ref{fig:lexic}).

This paper presents the following contributions:
\begin{itemize}[itemsep=0pt,topsep=2pt,leftmargin=10pt]
\item We create a novel dataset for Czech delexicalized generation; this extends the typical task of data-to-text NLG by requiring attribute value inflection (Section~\ref{sec:dataset}). We choose Czech as an example of a morphologically complex language \cite{cotterell_are_2018} with a large set of NLP tools readily available \cite[e.g.][]{popel_tectomt:_2010,strakova_open-source_2014,straka_tokenizing_2017}.
\item We present baseline models based on the TGen sequence-to-sequence (seq2seq) system \cite{dusek_sequence--sequence_2016}, with two novel extensions to the model for our task (Section~\ref{sec:model}):
\begin{itemize}[itemsep=0pt,topsep=0pt,leftmargin=12pt]
\item A model for lexicalization, i.e., selecting the correct inflected surface form for a slot value, based on a recurrent neural network language model (RNN LM);
\item A new generation mode, where the seq2seq generator produces interleaved sequences of lemmas (base word forms) and morphological tags that are postprocessed using a morphological generator.
\end{itemize}
\item Using both automatic and manual evaluation in Section~\ref{sec:experiments}, we show that our extensions improve over the base model, but do not solve the task completely.
\end{itemize}
We propose improvements for future work in Section~\ref{sec:conclusions}.
Our dataset and all experimental code are released on GitHub.\footnote{Dataset: \url{https://github.com/UFAL-DSG/cs_restaurant_dataset}, code: \url{https://github.com/UFAL-DSG/tgen}.}

\section{Dataset}
\label{sec:dataset}

Our goal was to create a dataset comparable in size and domain to existing English data-to-text NLG datasets used in experiments with neural systems. 
Since there are few to none Czech speakers on crowdsourcing platforms \cite{pavlick_language_2014,dusek_alex:_2014}, we were not able to use them for data collection.
Recruiting freelance translators seemed easier than training annotators; 
therefore, we turned to localizing and translating an existing dataset instead of creating a new one from scratch.
We chose 
the restaurant dataset of \citet{wen_semantically_2015} due to its manageable, yet non-trivial size and the familiarity of the domain \cite[cf.][]{mairesse_phrase-based_2010,dusek_evaluating_2019}. The original dataset contains 5,192 MR-sentence pairs, where MRs come in the form of dialogue acts (DAs).
A DA consists of DA type (e.g., \emph{request}, \emph{confirm}, \emph{inform}) and a list of slots (attributes) and their values (e.g., \emph{name}, \emph{price\_range}, \emph{address}, \emph{area}). There are 8 different DA types and 12 slots in the dataset.
All slots except the binary \emph{kids\_allowed} are delexicalized during generation (cf.~Figure~\ref{fig:lexic}).


\subsection{Localizing the Data}

\begin{figure}
\small
\begin{description}[leftmargin=!,labelwidth=1.7cm,itemsep=1pt]
\item[\emph{Ananta}\hfill\rm–] feminine noun, inflected (nom: \emph{Ananta}, gen: \emph{Ananty}, dat, loc: \emph{Anantě}, acc: \emph{Anantu}, inst: \emph{Anantou})
\item[\emph{BarBar}\hfill\rm–] masculine inanimate noun, inflected (nom, acc: \emph{BarBar}, gen, dat, loc: \emph{BarBaru}, inst: \emph{BarBarem})
\item[\emph{Café Savoy}\hfill\rm–] neuter noun, not inflected
\item[\emph{Místo}\hfill\rm–] neuter noun, inflected (nom, acc: \emph{Místo}, gen: \emph{Místa}, dat: \emph{Místu}, loc: \emph{Místě}, inst: \emph{Místem})
\item[\emph{U Konšelů}\hfill\rm–] prepositional phrase, not inflected
\end{description}
\caption{Examples of restaurant names from the localized data with different morphosyntactic behavior (\emph{nom} = nominative, \emph{gen} = genitive, \emph{dat} = dative, \emph{acc} = accusative, \emph{loc}~=~locative, \emph{inst} = instrumental).}
\label{fig:cs-infl-examples}
\end{figure}

We first needed to localize the dataset, replacing the original setting of San Francisco with a Czech one. 
In particular, we aimed at using domestic entity names (DA slot values) that need to be inflected since foreign names are often kept uninflected in Czech, using less fluent and conspicuous grammatical constructions to avoid inflection.\footnote{This is not to say that we avoided using any foreign words in the localization process. Since foreign restaurant names are quite common in Czechia, we also included some of them in the localized data.}

We localized the following slots in both DAs and texts from the dataset: restaurant names, areas, food types, street addresses, and landmarks.
We used a list of randomly chosen restaurant names from the Prague city center as well as lists of Prague neighborhoods, streets, and landmarks.
The resulting sentences contain mostly factually inaccurate, yet meaningful utterances about restaurants in Prague.

The localized lists are quite short, with just 15 different restaurant names and a similar number of landmarks, streets, and neighborhoods. While much longer lists would be needed for a real-world scenario, this is sufficient to cover most common classes of names with different inflection patterns and/or syntactic behavior (see Figure~\ref{fig:cs-infl-examples}). 

\subsection{Translation}

We recruited six translators and asked them to translate all unique texts in the localized dataset. 
They were given the following instructions:
\begin{itemize}[itemsep=0pt,topsep=2pt,leftmargin=10pt]
\item translate the utterances in isolation,
\item use fluent, spoken-style Czech,
\item strive to preserve the facts but not necessarily all nuances of the original, 
\item use varying synonyms (as long as they belong to casual, fluent Czech), including for entity names or slot values (such as price ranges or meal types),
\item inflect entity names as needed,
\item use formal address (or plural) when addressing the user, and use the female form in the first person for self-references.\footnote{Czech grammar requires a selection between formal an informal address whenever using a verb in the 2nd person \cite[p.~134ff.]{naughton_czech_2005}. For verbs with past tense or conditional and in any person, gender must be selected \cite[p.~140ff.]{naughton_czech_2005}. Here we opted for a feminine form whenever the system addresses itself, and formal address (mostly homonymous with plural) when addressing the user.}
\end{itemize}
All rules but the last one aim at obtaining a varied and fluent dataset; the last rule strives for consistency. Note that the translators were not given the input DAs -- these carry no more information than the corresponding English sentences, and we assume that they would only confuse the translators and could hurt the fluency of the results.


\subsection{Consistency Checks and Deduplication}


We checked the translated Czech texts for the presence of all required slot values. We took the following iterative, partially automatic approach:
\begin{enumerate}[itemsep=0pt,topsep=2pt,leftmargin=12pt]
\item Create a list of possible inflected surface forms for all slot values in the dataset. We used the morphological generator of \citet{strakova_open-source_2014} to inflect the surface forms automatically and manually checked for errors. \label{li:init-cs-postprocess}
\item Given a DA and a translated sentence, check (using an automatic script) that the sentence contains surface forms for all slots in the DA.\label{li:cs-postprocess-check}
\item Given a sentence found by the script to miss a value, check if it contains an alternative surface form not included in the list from Step~\ref{li:init-cs-postprocess}. If so, add this alternative surface form to the list.
\item If the translated sentence does not contain any mention of the DA value, fix the translation.
\item Repeat from Step~\ref{li:cs-postprocess-check} until there are no missing DA value mentions in the whole set.
\end{enumerate}
Note that these checks result not only in greater consistency of the dataset, but also in a list of possible surface realizations for all slot values in the dataset. We store this list including morphological information provided by the tagger (with manually corrected errors), and we use it for lexicalization (see Section~\ref{sec:model}).\footnote{We treat multiword slot values as single tokens in our surface form list. We assign them a morphological tag that fits the whole expression best, e.g., a noun tag for noun phrases.}

\subsection{Duplicate Sentence Handling}
\label{sec:expansion}

\begin{figure*}[tb]
{\small
\glll mít pro ty vhodný restaurace\hspace{-1mm} .\hspace{-1mm} jeho název být\hspace{-2mm} X-name a moci se dát X-food kuchyně\hspace{-1mm} .
Mám pro Vás vhodnou restauraci\hspace{-1mm} .\hspace{-1mm} Její název je {Kočár z Vídně} a můžete si dát českou kuchyni\hspace{-1mm} .
{I have\hspace{-2mm}} for you {a suitable\hspace{-2mm}} restaurant\hspace{-1mm} .\hspace{-1mm} Its name is {Kočár z Vídně} and {you can} yourself\hspace{-2mm} give Czech cuisine\hspace{-1mm} .
\gln 
\glend

‘I have a suitable restaurant for you. Its name is Kočár z Vídně and you can have Czech cuisine.’
}
\caption{Lemmatized and delexicalized form of the translations for LM scoring.
Top: lemmatized and delexicalized Czech used for the LM; middle: original Czech sentence including lexicalization; bottom: English word-by-word gloss. An English translation is shown below the example.
}
\label{fig:expand-lm}
\end{figure*}

If the exact lexicalization is not taken into account, the original dataset of \citet{wen_semantically_2015} contains a lot of duplicate texts -- the total number of DA-text pairs is 5,192, but only 2,648 are unique. Therefore, we chose to only translate unique texts, in order to speed up the translation process and lower the costs, albeit at a cost of a lower-quality result. 
We ensured that the translations preserve the same number of unique sentences by modifying any duplicate translations, manually replacing selected words or phrases with synonyms.

After the dataset was translated, we expanded it
to obtain the same number of instances and the same distribution of different DAs as in the original. 
Given a delexicalized DA, a list of corresponding translated sentences, and the target number of corresponding sentences to match the original set, we sampled additional copies of the existing translations to match the number of originals.
To estimate probabilities of the individual translations for the sampling, we used a 5-gram LM\footnote{We used the implementation in the KenLM toolkit \cite{heafield_kenlm:_2011}.} trained on lemmatized and delexicalized translations (see Figure~\ref{fig:expand-lm} for details).  We obtained LM scores for all translations, used  softmax to obtain a probability distribution, and sampled additional copies from this distribution.
This ensures that translations using more frequent phrasing are more likely to be used multiple times in the set.

We then relexicalized the sampled copies: We randomly changed DA slot values and replaced their surface forms in the text using the surface forms list, checking for roughly corresponding morphology.
Since the morphological information used by this approach was rather crude (e.g., noun/adjective gender was not taken into account), disfluencies ensued in some cases. Therefore, we manually corrected all relexicalized sentences, changing inflection or wording where needed. 

\subsection{Dataset Statistics}

\begin{table}[tb]
\begin{center}\small
\begin{tabular}{lrr}\hline
                               & \bf English  & \bf Czech\Tstrut\Bstrut \\\hline
Number of instances            &           5,192      & 5,192\Tstrut \\
Unique delexicalized instances &           2,648      & 2,752 \\
Unique delexicalized DAs       &             248      &   248 \\
Unique lemmas  (in delexicalized set)\hspace{-1cm}      &  399     &   532 \\
Unique word forms   (in delexicalized set)\hspace{-0.5cm} &  455     &   962 \\
Average lexicalizations per slot value\hspace{-1cm}     &   1      &   3.84 \\\hline
\end{tabular}
\end{center}
\caption{Statistics of our translated Czech dataset and a comparison to the English original of \citet{wen_semantically_2015}.
The average lexicalizations per slot value shows the number of different surface lexical forms per slot value, as it appears in the dataset. Numerals were disregarded when computing this value. 
}
\label{tab:cs-set-stats}
\end{table}

The final Czech set contains the same number of instances as the English original, copies the DA distribution of the original, and contains a slightly higher number of unique delexicalized sentences due to post-expansion corrections (see Section~\ref{sec:expansion}).
A statistics of the dataset size is shown in Table~\ref{tab:cs-set-stats}, with a comparison to the original English set. We can see that while the number of unique word lemmas (disregarding restaurant and place names) is slightly higher in the Czech set, the number of unique inflected word forms is more than twice as high.
It is also clear that using slot values verbatim in the text is not possible in the Czech set as the number of possible lexical realizations for each value is much higher than one.


\subsection{Data Split}
\label{sec:split}

The original dataset of \citet{wen_semantically_2015}, which used a sequential 3:1:1 split into training, development and test parts, suffered from a lot of overlap in terms of delexicalized DAs between the sections. This means that a system can perform quite well on this dataset and still be unable to generalize to unseen DAs \cite{lampouras_imitation_2016}.
To make testing systems' generalization capabilities possible on our Czech dataset, we opted for a different data split. We roughly keep the same 3:1:1 size proportion (see Table~\ref{tab:split}), but we make sure no delexicalized DA appears in two different parts. On the other hand, we ensure that most DA types (\emph{inform}, \emph{confirm} etc.) are represented in all data parts, so the system has access to all general types of sentences during training.\footnote{This is impossible to achieve for the \emph{goodbye} and \emph{?reqmore} DA types (i.e., goodbyes and asking if the user needs anything else). These DA types never appear with slots and thus only have one corresponding DA. We keep the corresponding instances in the training set.}

\begin{table}[tb]
\centering\small
\begin{tabular}{lrrr}\hline
Part                      &  \bf Train & \bf Dev & \bf Test\Tstrut\Bstrut \\\hline
Unique delexicalized DAs  &  144      & 51          & 53 \\
Total number of instances & 3,569 & 781 & 842 \\\hline
\end{tabular}
\caption{Dataset split statistics.}
\label{tab:split}
\end{table}

\section{Model}
\label{sec:model}

We use TGen \cite{dusek_sequence--sequence_2016} in our experiments, which is a freely available NLG system based on the seq2seq model with attention \cite{bahdanau_neural_2015}.

The seq2seq model consists of the encoder, the decoder, and the attention model. Both the encoder and decoder are recurrent neural networks (RNN) with LSTM cells \cite{hochreiter_long_1997}. The encoder takes the input DA as a sequence of triples ``DA type -- slot -- value''\footnote{DA type is repeated for each slot-value pair.} and produces a sequence of hidden states. The last hidden state is used to initialize the decoder, all hidden states serve as input into the attention model. The attention model produces their weighted combination for each decoder step using a 1-layer fully connected network. The decoder generates output tokens one-by-one using the previously generated token and the attention model as inputs.

In addition to the basic seq2seq model, TGen adds beam search and a reranker for the candidate outputs on the generation beam that checks if the input semantics is preserved. The reranker encodes a candidate output using an LSTM RNN and produces a binary classification of DA types and slot-value pairs present. The number of differences against the input DA is used as penalty.


\subsection{Basic Extensions}

We added two features fairly standard in seq2seq-based models but absent from TGen:
\begin{itemize}[itemsep=0pt,topsep=2pt,leftmargin=10pt]
\item Bidirectional encoder \cite{bahdanau_neural_2015} -- the input sequence is encoded in both directions and the resulting hidden states are joined. We added this for both the main seq2seq generator and the reranker.
\item Dropout \cite{hinton_improving_2012} -- this zeroes out certain connections within the network with a given probability during the training process; it serves as regularization feature. We use this in the main generator only.    
\end{itemize}
We use these extensions in all our setups as they improved results in our preliminary experiments.

\subsection{Lemma-tag Generation Mode}
\label{sec:lemma-tag}

\begin{figure*}[tb]
\small\raggedright
{\gll hledat {\tt VB-P---2P-AA---} vhodný {\tt AAFS4----1A----} restaurace {\tt NNFS4-----A----} 
search {verb, 2nd person present formal} suitable {adjective, fem~sg~acc} restaurant {noun, fem~sg~acc}
\gln\glend}
{\gll na {\tt RR--4----------} X-good\_for\_meal {\tt NNFS4-----A----} ? {\tt Z:-------------}
for {preposition, acc} {slot placeholder} {noun, fem~sg~acc} ? {final punctuation}
\gln\glend}
\caption{Example interleaved lemma-tag sequence for the input DA \emph{?confirm(good\_for\_meal=breakfast)}, the first output from Figure~\ref{fig:lexic} (acc = accusative, fem = feminine, sg = singular; cf.~\cite{hajic_disambiguation_2004} for tagset details). Note that the morphological tag for the slot placeholder is included and can be used during lexicalization (cf.~Section~\ref{sec:lexicalization}).}
\label{fig:interleaved}
\end{figure*}

\citet{dusek_sequence--sequence_2016} experiment with generating syntactic trees and realizing them using an external surface realizer; they report slightly worse performance than generating tokens directly. 

In order to fight data sparsity coming from the rich morphology of Czech,
we decided to explore the middle ground between syntactic trees and full word-form generation: generating base forms (lemmas) and morphological tags that indicate how the form should be inflected. 
We train TGen to simply generate an interleaved sequence of lemmas and tags (see Figure~\ref{fig:interleaved}), which are then postprocessed using the dictionary-based morphological generator of \citet{strakova_open-source_2014} to obtain the inflected word forms.

In the lemma-tag mode, the set of possible output tokens is reduced compared to direct token generation, but the postprocessing step is much simpler than using a full syntactic surface realizer.
Moreover, the generated morphological tags following slot placeholders can be used to limit the scope of possible surface forms during lexicalization (see Section~\ref{sec:lexicalization}).

This approach is inspired by similar approaches in phrase-based MT \cite{bojar_english--czech_2007,toutanova_applying_2008,fraser_experiments_2009}
and was developed in parallel to recent similar experiments with two-step neural MT \cite{nadejde_predicting_2017,tamchyna_modeling_2017}.
We compare the lemma-tag generation mode against the TGen default direct word-form generation mode in our experiments.

\subsection{Lexicalization}
\label{sec:lexicalization}

We experiment with three different approaches for selecting the surface form for a DA slot value placeholder from a set of applicable ones -- two very straightforward baselines requiring no training and our proposed solution based on a neural LM:
\begin{itemize}[itemsep=0pt,topsep=2pt,leftmargin=10pt]
\item \emph{Random baseline.} This selects a surface form at random. This approach is certainly not suitable for a real application, we only use it for comparison.

\item \emph{Most frequent baseline.} Here, the applicable surface form that occurs overall most frequently in the training data is selected. This represents a stronger baseline than the random method.

\item \emph{RNN-based language model.} Our main solution attempts to choose the best surface form using a bidirectional LSTM RNN-based LM \cite{mikolov_recurrent_2010}, 
trained to predict a token probability distribution given all previous and all following tokens.
During decoding, the RNN LM estimates the probabilities of all applicable surface forms, and we select the most probable surface form for the output.
\end{itemize}

When selecting a surface form during direct word-form generation, all possible forms for the given slot value are considered.
In the lemma-tag mode (Section~\ref{sec:lemma-tag}), only forms matching the morphological tag following the slot placeholder are considered (cf.~Figure~\ref{fig:interleaved}) -- first the ones matching perfectly, with backoffs to coarse part-of-speech or all possible forms.

\subsection{Lexicalized Input DAs}
\label{sec:lexicalized-das}

Some slot values in our dataset may require certain morphosyntactic structure of their neighborhood. This is the case for restaurant counts: Czech cardinal numerals 2-4 behave as adjectives, while higher numerals behave as nouns and take the counted quantity as genitive object. The correct nominative forms when counting restaurants are then ``2 restaurac\emph{e}'', but ``5 restaurac\emph{í}'' \cite[p.~113ff.]{naughton_czech_2005}. Another example are area names requiring different prepositions for location -- the correct form for ``in Malá Strana'' is ``\emph{na} Malé Straně'', but for ``in Karlín'', it is ``\emph{v}~Karlíně'' \cite[p.~202]{naughton_czech_2005}.

Therefore, inspired by \citet{sharma_natural_2017}, we test using fully lexicalized input DAs with the main generator to check if it learns to produce more appropriate structure for concrete values (while still producing delexicalized output).\footnote{We exploit the fact that the number of possible values for different slots in the dataset is relatively small (cf.~Section~\ref{sec:dataset}); morphosyntactic classes of the values would need to be used if the number of values was higher.}
We compare this setup against the default with delexicalized DAs.

\section{Experiments}
\label{sec:experiments}

\subsection{Experimental Setup}

We test all combinations of the features described in Section~\ref{sec:model}:
\begin{itemize}[itemsep=0pt,topsep=2pt,leftmargin=10pt]
    \item Direct token vs.\ lemma-tag generation
    \item Random / most-frequent / RNN LM lexicalizer
    \item Delexicalized vs.\ lexicalized input DAs
\end{itemize}

We train the resulting 12 model variants using the Adam optimizer \cite{kingma_adam:_2015} to minimize cross entropy on the training set; this approach is used for all parts of the system: the main seq2seq generator, the reranker, and the RNN LM lexicalizer. After each training data pass, we validate the models and keep the best-perform\-ing parameters. We use BLEU score \cite{papineni_bleu:_2002}, classification error, and LM perplexity as the respective validation criteria. We set hyperparameters based on TGen defaults for other datasets and a few experiments on the development set.\footnote{
\label{fn:params}The main generator uses embedding and LSTM cell size 200, learning rate 0.005, dropout rate 0.5, and batch size 20. At least 50 and up to 1000 training data passes are used, with early stopping if the top 10 validation BLEU scores do not change for 50 passes. Beam size 20 is used for decoding.

The reranker uses embedding and LSTM cell size 50, no dropout, learning rate 0.001, and batch size 20. Training runs for 100 passes, performance is validated starting with pass 10. The reranker is validated both on training and development data; classification error on the development set is given 10 times more weight than training set error.

The RNN LM lexicalizer uses the same parameters as the reranker, with training for 50 passes maximum and validation (on development data only) starting after the first pass.
}

Training the baseline lexicalizers is trivial: the random baseline does not require any training, it simply uses the list of possible surface forms; the most frequent baseline just memorizes surface form frequencies in the training data.

To reduce the effect of random initialization, we train five runs using different random seeds and use results of all of them for evaluation. In addition, we fix the random seeds so that identical seq2seq generators and rerankers are used in setups that only differ in the lexicalization method.

\begin{table*}[tb]
\centering\small
\begin{tabular}{lllllcccc}\hline
\bf Input DAs & \bf Generator mode & \bf Lexicalizer & \bf BLEU & \bf NIST & \bf\hspace{-3mm}METEOR\hspace{-1mm} & \bf\hspace{-1mm}ROUGE-L\hspace{-2mm} & \bf CIDEr & \bf SER  \\\hline
\multirow{6}{*}{Delexicalized} & \multirow{3}{*}{Word forms} 
        & Random        & 15.51$^\ddag$  & 3.7352  & 18.60  & 35.00  & 1.3922  & \bf 00.70    \\ 
    &   & Most frequent & 20.28$^\ddag$  & 4.5192  & 22.69  & 40.92  & 1.9399  & \bf 00.70    \\ 
    &   & RNN LM        & 20.74$^\ast$  & 4.5096  & 22.61  & 40.72  & 1.9924  & \bf 00.70    \\\cdashline{2-9}[0.5pt/2pt] 
                               & \multirow{3}{*}{Lemma-tag}  
        & Random        & 19.66$^\dag$  & 4.4884$^{\dag\ddag}$  & 22.19  & 41.42  & 1.8844  & 01.85    \\ 
    &   & Most frequent & 21.21$^{\dag\ddag}$  & 4.6900$^{\dag\ddag}$  & 23.07  & 42.62  & 2.0983  & 01.85    \\ 
    &   & RNN LM        & \bf 21.96$^{\ast\dag\ddag}$  & \bf 4.7720$^{\ast\dag\ddag}$  & \bf 23.32  & \bf 42.95  & \bf 2.1783  & 01.85    \\\hline 
\multirow{6}{*}{Lexicalized}   & \multirow{3}{*}{Word forms} 
        & Random        & 14.70  & 3.7595  & 18.29  & 35.64  & 1.3712  & 02.30    \\ 
    &   & Most frequent & 19.73  & 4.5618  & 22.45  & 41.71  & 1.9473  & 02.30    \\ 
    &   & RNN LM        & 20.48$^\ast$  & 4.6060$^{\ast\ddag}$  & 22.55  & 41.66  & 2.0192  & 02.30    \\\cdashline{2-9}[0.5pt/2pt] 
                               & \multirow{3}{*}{Lemma-tag}  
        & Random        & 18.92$^\dag$  & 4.3501$^\dag$  & 21.76  & 40.55  & 1.8014  & 03.08    \\ 
    &   & Most frequent & 19.44  & 4.4453  & 22.22  & 41.26  & 1.8801  & 03.08    \\ 
    &   & RNN LM        & 20.42$^\ast$  & 4.5460$^\ast$  & 22.56  & 41.73  & 1.9796  & 03.08    \\\hline 
\end{tabular}
\caption{Automatic metrics results. See Section~\ref{sec:metrics} for metrics; scores are averaged over 5 different random initializations, all scores except for NIST and CIDEr are percentages. $^\ast$ = significantly better than the corresponding most frequent baseline lexicalizer, $^\dag$ = significantly better than the corresponding word forms mode, $^\ddag$ = significantly better than the corresponding (de)lexicalized input DAs. Significance was assessed using pairwise bootstrap resampling \cite{koehn_statistical_2004}, $p<0.01$.}
\label{tab:autom-results}
\end{table*}

\begin{table*}[tb]
\centering\small
\begin{tabular}{lllrrrrrcc}\hline
\bf Input DAs & \bf Generator mode & \bf Lexicalizer & \bf S & \bf R & \bf F & \bf I & \bf L & \bf F+I+L & \bf $\mathbf{\Sigma}$  \\\hline
\multirow{4}{*}{Delexicalized}  & \multirow{2}{*}{Word forms} 
        & Most frequent &  \bf 8 &   \bf 0 &  \bf 5 &  11 &  57 &  73 &  81 \\ 
    &   & RNN LM        &  \bf 8 &   \bf 0 &  \bf 5 &  11 &  25 &  41 &  49 \\\cdashline{2-10}[0.5pt/2pt] 
                                & \multirow{3}{*}{Lemma-tag}  
        & Most frequent & 12 &   2 &  \bf 5 &  11 &  45 &  61 &  75 \\ 
    &   & RNN LM        & 12 &   2 &  \bf 5 &  11 &   6 &  22 &  36 \\\hline 
\multirow{4}{*}{Lexicalized}   & \multirow{2}{*}{Word forms} 
        & Most frequent & 14 &   5 &  14 &   6 &  34 &  54 &  73 \\ 
    &   & RNN LM        & 14 &   5 &  14 &   6 &  10 &  30 &  49 \\\cdashline{2-10}[0.5pt/2pt] 
                                & \multirow{3}{*}{Lemma-tag}  
        & Most frequent & 15 &   4 &   6 &   \bf 4 &  34 &  44 &  63 \\ 
    &   & RNN LM        & 15 &   4 &   6 &   \bf 4 &   \bf 4 &  \bf 14 &  \bf 33 \\\hline 
\end{tabular}
\caption{Manual evaluation results on 100 sampled sentences -- absolute numbers of different types of errors (S = semantic errors, R = repetition, F = fluency problems except lexicalization, I = impossible to lexicalize correctly with the given value, L = lexicalization errors). All error types are exemplified in Figure~\ref{fig:example-outputs}.}
\label{tab:human-results}
\end{table*}

\subsection{Metrics}
\label{sec:metrics}

We use the suite of word-overlap-based automatic metrics from the E2E NLG Challenge \cite{dusek_evaluating_2019},\footnote{\url{https://github.com/tuetschek/e2e-metrics}} supporting BLEU \cite{papineni_bleu:_2002}, NIST \cite{doddington_automatic_2002}, ROUGE-L \cite{lin_rouge:_2004}, METEOR \cite{lavie_meteor:_2007} and CIDEr \cite{vedantam_cider:_2015}.
Although multiple texts often correspond to the same delexicalized DA, we treat each instance individually both in training and testing since the particular slot values influence the shape of the whole sentence (see Sections~\ref{sec:expansion} and~\ref{sec:lexicalized-das}). This means that only a single reference output per instance is available to be used with automatic metrics (see Section~\ref{sec:results}).

In addition to word-overlap metrics, we use the slot error rate \cite[SER;][]{wen_semantically_2015} to evaluate semantic accuracy of the outputs. This metric counts slot placeholders in the output before lexicalization and compares them to slots in the input DA. It reliably measures the amount of missed/added content in all delexicalized slots (cf.~Section~\ref{sec:dataset}), but the non-delexicalized binary \emph{kids\_allowed} slot is ignored.

\subsection{Results}
\label{sec:results}

The automatic metrics scores for all setups are shown in Table~\ref{tab:autom-results}. In terms of generator mode, using lemma-tag generation significantly\footnote{BLEU and NIST differences are statistically significant ($p<0.01$) according to bootstrap resampling \cite{koehn_statistical_2004}.\label{fn:bootstrap}} improves word-overlap metrics over direct token generation in the delexicalized input setting. However, it also leads to an increased SER.
The RNN LM brings a significant\textsuperscript{\ref{fn:bootstrap}} improvement over both baselines in all setups; the very low performance of the random baseline only documents that inflection indeed matters for slot values.
The lexicalized input DAs did not bring improvement over the delexicalized setting -- lexicalized setups seem to perform slightly worse in terms of both word-overlap metrics and SER.


\subsection{Manual Error Analysis}

\begin{figure*}[tb]
\centering\scriptsize
\begin{tabular}{lll}
\multicolumn{2}{l}{\bf Input DA}   & \emph{inform(food=Turkish, name=``Green Spirit'', price\_range=expensive)}\hspace{5cm} \\
\parbox[t]{1mm}{\multirow{14}{*}{\rotatebox[origin=c]{90}{\bf Systems}}} & \stack{Delex. input DAs, any mode\\Most frequent} & \TTstrut{}Green Spirit \g{je}{is} \g{\lexic{drahé}}{expensive} \g{turecká}{Turkish} \g{restaurace}{restaurant}.\\
&\stack{Delex. input DAs, any mode\\RNN LM} & \TTstrut{}Green Spirit \g{je}{is} \g{drahá}{expensive} \g{turecká}{Turkish} \g{restaurace}{restaurant}.\\
&\stack{Lex. DAs, word forms\\Most frequent} & \TTstrut{}Green Spirit \g{je}{is} \g{\lexic{drahé}}{expensive} \g{turecká}{Turkish} \g{restaurace}{restaurant} \\
&\stack{Lex. DAs, word forms\\RNN LM} & \TTstrut{}Green Spirit \g{je}{is} \g{drahá}{expensive} \g{turecká}{Turkish} \g{restaurace}{restaurant} \\
&\stack{Lex. DAs, lemma-tag\\Most frequent} & \TTstrut{}Green Spirit je \g{\lexic{drahé}}{expensive} \g{restaurace}{restaurant}, \g{která}{which} \g{podává}{serves} \g{turecká}{Turkish} \g{jídla}{meals}.\\
&\stack{Lex. DAs, lemma-tag\\RNN LM} & \TTstrut{}\BBstrut{}Green Spirit \g{je}{is} \g{drahá}{expensive} \g{restaurace}{restaurant}, \g{která}{which} \g{podává}{serves} \g{turecká}{Turkish} \g{jídla}{meals}.\\
\end{tabular}
\begin{tabular}{lll}
\multicolumn{2}{l}{\bf Input DA}   & \TTstrut{}\emph{inform(area=dont\_care, count=218, food=dont\_care, price\_range=dont\_care, type=restaurant)} \\
\parbox[t]{1mm}{\multirow{10}{*}{\rotatebox[origin=c]{90}{\bf Systems \rm(any lexicalizer)}}} & \TTstrut{}\stack{Delex.~input DAs\\Word forms mode} & \g{Pokud}{if} \g{vám}{you} \g{nezáleží}{don't-care} \g{na}{about} \g{druhu}{type} \g{jídla}{of-food}, \g{našla jsem}{I-found} 218 \g{\inapstr{restaurace}}{restaurants} \g{v}{in} \g{různých}{various} \g{cenových}{price} \g{skupinách}{ranges}. \semerr{(missing area)} \\
& \TTstrut{}\stack{Delex.~input DAs\\Lemma-tag mode} & \g{Pokud}{if} \g{vám}{you} \g{nezáleží}{don't-care} \g{na}{about} \g{druhu}{type} \g{jídla}{of-food}, \g{našla jsem}{I-found} 218 \g{restaurací}{restaurants} \g{v}{in} \g{různých}{various} \g{cenových}{price} \g{skupinách}{ranges}. \semerr{(missing area)} \\
& \multirow{2}{*}{\stack{Lex.~input DAs\\Word forms mode}} & \TTstrut{}\g{V}{into} \g{úvahu}{consideration} \g{\inapstr{připadají}}{come} 218 \g{restaurací}{restaurants}, \g{pokud}{if} \g{vám}{you} \g{nezáleží}{don't-care} \g{na}{about} \g{druhu}{type} \g{jídla}{of-food}\repeat{,} \g{\repeat{pokud}}{if} \g{\repeat{vám}}{you} \g{\repeat{nezáleží}}{don't-care} \g{\repeat{na}}{about} \g{\repeat{druhu}}{type} \g{\repeat{jídla}}{of-food}. \\
& & \semerr{(missing area, price range)}\\
& \TTstrut{}\stack{Lex.~input DAs\\Lemma-tag mode} & \g{Mám}{I-have} \g{tu}{here} 218 \g{restaurací}{restaurants}, \g{pokud}{if} \g{vám}{you} \g{nezáleží}{don't-care} \g{na}{about} \g{\fluency{druhu}}{type} \g{cenových}{price} \g{skupinách}{ranges}. \semerr{(missing area, food type)} \\
\end{tabular}
\caption{Examples from manual error analysis. Errors are marked with color and underlining: \semerr{semantic errors}, \repeat{repetition}, \fluency{fluency}, \inapstr{impossible to lexicalize correctly}, \lexic{lexicalization} (cf.~Table~\ref{tab:human-results}). In the top example, the RNN LM lexicalizer is able to select the correct feminine singular form, while the most frequent baseline selects a neuter form. In the bottom example, systems with lexicalized input DAs make more semantic errors. The lemma-tag mode is able to select a more appropriate syntactic structure for the numeral 218.}
\label{fig:example-outputs}
\end{figure*}

To obtain a deeper insight into the results and account for automatic metrics' inaccuracy \cite{novikova_why_2017,reiter_structured_2018}, we performed a detailed manual error analysis on a sample of 100 outputs produced by all systems except the ones with random baseline lexicalizers, which clearly perform poorly. This was a blind annotation of semantic and fluency errors; it is not a preference rating. We categorized multiple error types; the results are shown in Table~\ref{tab:human-results}.

The analysis confirmed that lexicalized input DAs cause more semantic errors (both missed slots and repetition). On the other hand, the outputs were more fluent in this setting, which is not apparent with automatic metrics.
Lemma-tag generation also improves fluency overall, at the cost of increasing the number of semantic errors.
The RNN LM lexicalizer leads to significant reduction of lexicalization errors compared to the most frequent baseline, especially in combination with lemma-tag generation (see top example in Figure~\ref{fig:example-outputs}).
None of the systems produce perfect output; they seem to struggle especially with DAs that are very different from the ones found in the training set and/or occur less frequently (see bottom example in Figure~\ref{fig:example-outputs}, cf.~Section~\ref{sec:split}).
We believe that an increased amount of training data could improve the situation.

\section{Related Work}
\label{sec:related}

NLG experiments for non-English languages are relatively rare and fully trainable approaches even rarer. Our work is, to our knowledge, the first application of neural NLG to a non-English language for data-to-text generation.

Most works concerned with multiple languages focus on surface realization.
There have been a few approaches using handcrafted grammars 
\cite{bateman_enabling_1997,allman_linguists_2012}.
The procedural SimpleNLG realizer \cite{gatt_simplenlg:_2009} has also been ported into multiple languages \cite{bollmann_adapting_2011,vaudry_adapting_2013,de_oliveira_adapting_2014,mazzei_simplenlg-it:_2016,ramos-soto_adapting_2017,cascallar-fuentes_adapting_2018,chen_simplenlg-zh:_2018,de_jong_going_2018}.
Further works using multilingual rule-based surface realization pipelines were developed in the context of machine translation \cite{aikawa_generation_2001,zabokrtsky_tectomt:_2008,dusek_new_2015}.
\citet{bohnet_broad_2010} created the first statistical multilingual realizer based on a pipeline of SVMs, 
the recent surface realization challenge \cite{mille_first_2018} then features further fully trainable realizers tested on multiple languages, including neural models.

In data-to-text generation, the recent work of \citet{moussallem_rdf2pt:_2018} is applied to Portuguese, but is largely rule-based.
The works of \citet{chen_training_2010} and \citet{kim_generative_2010} represent the only data-to-text end-to-end NLG system with multilingual experiments known to us; they generate English and Korean sport commentary sentences using an inverted (non-neural) semantic parser. Our dataset is ca.~2.5 times larger and more complex, given the slot value inflection.

Other works on neural non-English NLG solve in fact different tasks from ours: Chinese poetry generation \cite{zhang_chinese_2014,yi_generating_2016,wang_chinese_2016}, non-task-oriented response generation in chatbots \cite{xing_topic_2016,xing_topic_2017}, or morphological inflection \cite[e.g.][]{faruqui_morphological_2016,kann_single-model_2016}.

\section{Conclusions and Future Work}
\label{sec:conclusions}

We presented the first dataset targeted at end-to-end neural non-English NLG, containing Czech texts from the restaurant domain. 
We show that the task of data-to-text NLG here is harder as slot values require morphological inflection.
We apply  to our data the freely available, state-of-the-art TGen NLG system \cite{dusek_sequence--sequence_2016} based on the seq2seq architecture, and we implement two extensions for  Czech: 
(1) an RNN LM model to select the correct inflected surface form for slot values and (2) lemma-tag generation mode, where the generator produces an interleaved sequences of base form and morphological tags, which are postprocessed by a morphological generator.
We also experiment with lexicalized and delexicalized slot values in generator inputs.
Using both automatic metrics and manual analysis, we show that the RNN LM brings clear benefits. The lemma-tag mode and lexicalized inputs improve fluency but hurt semantic accuracy of the outputs.
We release our dataset dataset and all experimental code on GitHub.\footnote{Dataset: \url{https://github.com/UFAL-DSG/cs_restaurant_dataset}, code: \url{https://github.com/UFAL-DSG/tgen}.}

In future work, we will collect a large unannotated dataset and pretrain the generator \cite{chen_few-shot_2019}. We believe that this will lead to increased output fluency and accuracy. We are also considering using machine translation to obtain more synthetic training data points.

\section*{Acknowledgments}

This research was supported by the Charles University project PRIMUS/19/SCI/10 and by the Ministry of Education, Youth and Sports of the Czech Republic under the grant agreement LK11221. 
This work used using language resources distributed by the LINDAT/CLARIN project of the Ministry of Education, Youth and Sports of the Czech Republic (project LM2015071).

\bibliography{references}
\bibliographystyle{acl_natbib}

\end{document}